\documentclass[graybox]{svmult}

\usepackage[T1]{fontenc}
\usepackage{url}
\usepackage{graphicx} 
\usepackage{subfig}

\begin{document}

\title*{Simulation of microstructures and machine learning}
\author{Katja Schladitz and Claudia Redenbach and Tin Barisin and
Christian Jung and Natascha Jeziorski and Lovro Bosnar and
Juraj Fulir and Petra Gospodneti{\'c}}
\authorrunning{K. Schladitz, C. Redenbach, T. Barisin, J. Fulir, P. Gospodneti{\'c} et al} 

\institute{Katja Schladitz \at Fraunhofer ITWM, Image Processing, Kaiserslautern, \email{katja.schladitz@itwm.fraunhofer.de}
\and Claudia Redenbach \at RPTU, Department of Mathematics, Kaiserslautern, \email{claudia.redenbach@rptu.de} 
\and Tin Barisin \at RPTU, Department of Mathematics
\and Christian Jung \at RPTU, Department of Mathematics, Kaiserslautern, \email{christian.jung@rptu.de}
\and Natascha Jeziorski \at RPTU, Department of Mathematics, Kaiserslautern, \email{n.jeziorski@rptu.de}
\and Lovro Bosnar \at RPTU, Department of Computer Science
\and Juraj Fulir \at Fraunhofer ITWM, Image Processing, Kaiserslautern, \email{juraj.fulir@itwm.fraunhofer.de} 
\and Petra Gospodneti{\'c} \at Fraunhofer ITWM, Image Processing, Kaiserslautern, \email{petra.gospodnetic@itwm.fraunhofer.de}
}

\maketitle

\abstract{Machine learning offers attractive solutions to challenging image processing tasks. 
Tedious development and parametrization of algorithmic solutions can be replaced by 
training a convolutional neural network or a random forest with a high potential to 
generalize. However, machine learning methods rely on huge amounts of representative 
image data along with a ground truth, usually obtained by manual annotation. 
Thus, limited availability of training data is a critical
bottleneck. We discuss two use cases: 
optical quality control in industrial production and segmenting crack structures in 3D images of concrete.
For optical quality control, all defect types have to be trained but
are typically not evenly 
represented in the training data. Additionally, manual annotation is 
costly and often inconsistent. 
It is nearly
impossible in the second case: segmentation of 
crack systems in 3D images of concrete. 
Synthetic images, generated based on realizations of stochastic geometry models, 
offer an elegant way out. A wide variety of structure types can be generated. 
The within structure variation is naturally captured by the stochastic nature of 
the models and the ground truth is for free. Many new questions
arise. In particular, which characteristics of the real image data have to be 
met to which degree of fidelity.
\footnote{This is a preprint of the following chapter: 
Katja Schladitz, Claudia Redenbach, Tin Barisin, Christian Jung,
Natascha Jeziorski, Lovro Bosnar, Juraj Fulir, and  Petra
Gospodneti{\'c}: Simulation of Microstructures and Machine Learning,
published in Continuum Models and Discrete Systems, edited by François
Willot, Justin Dirrenberger, Samuel Forest, Dominique Jeulin, Andrej
V. Cherkaev, 2024, Springer Cham, reproduced with permission of
Springer Nature Switzerland AG 2024. The final authenticated version
is available online at: 
https://doi.org/10.1007/978-3-031-58665-1}
}

\section{Introduction}

Machine learning offers attractive solutions to challenging image 
processing tasks. Tedious development and parametrization of algorithmic 
solutions can be replaced by training a convolutional neural network (CNN) 
or a random forest. Pre-trained CNN architectures are easily available.
Moreover, machine learning methods have higher potential to generalize 
than classical, tailor-suited solutions. However, ML methods rely on huge 
amounts of representative image data along with a ground truth, usually 
obtained by manual annotation. Limited availability of training data is 
therefore a critical bottleneck in a variety of image processing problems. 

Here, we discuss three cases: reconstruction of highly porous spatial 
structures from stacks of scanning electron microscopy (SEM) images of 
samples serially sectioned by a focused ion beam (FIB), segmenting crack 
structures in 3D images of concrete, and optical quality control in 
industrial production.

We start off with the reconstruction of highly porous structures 
from FIB-SEM stacks. This is a semantic segmentation task 
rendered exceptionally demanding by the high depth of view of the  
SEM leading to so-called shine-through-artifacts in the 3D image.
This challenge is a rather old example of successful use of synthetic 
image data derived from stochastic geometry models. FIB-SEM is an
established imaging method for materials' structures at the 
nano-scale \cite{HOLZER2016314}. However, deriving a truly spatial 
representation of the solid component of a highly porous material from
a stack of SEM images of consecutive slices involves a tedious 
segmentation (or reconstruction) step. Several solutions have been 
suggested \cite{prill13:jmi,SALZER201436}, but tailor-made 
algorithms like \cite{ott-roldan2019} prevail. Adaption to new structures 
or materials is still challenging. Synthetic training data can be 
generated \cite{prill12:scanning} and have been used successfully to 
train a 3d U-net \cite{fend2020}. Generalization to unseen data remains 
however a challenge.

The second use case is motivated by the rise of Gulliver at RPTU in 
Kaiserslautern -- a dedicated CT device for observing concrete beams under 
load \cite{gulliver, salamon:gulliver}. 
Concrete beams of up to 6\,m length will be imaged there spatially 
while undergoing bending tests. Gulliver will generate up to 
$10\,000\times 10\,000\times 2\,000$ voxels per scan and several scans per 
in-situ bending experiment. Consistent manual annotation of cracks in CT images 
of concrete is particularly challenging if not even impossible due to the 
cracks being very thin and dark. Moreover, for the Gulliver generated images, 
interactive crack detection and segmentation is completely out of reach due to 
the mere amount of data. Hence, algorithms for automatic crack segmentation 
have been collected from previous work \cite{paetsch11, paetsch12}, modified, 
and newly developed \cite{barisin22}. For a fair quantitative comparison as 
well as for the training of ML models, CT images of cracked concrete samples 
with a ground truth were needed. Simulated crack structures 
\cite{jung22cracks} impressed on real CT images of concrete samples yield both 
-- a ground truth and a degree of realism allowing for generalization to real 
crack structures. Additional effort is however needed to adapt to the strong 
variation in thickness of real crack structures \cite{ndt-ce,ict22}. The scale 
invariant RieszNet \cite{barisin22riesz} is the most innovative answer to this 
challenge.

The third case focuses only on the 2D imaging of the object's surface, using 
non-penetrating light. It is a popular research field due to its relatively low 
cost of data acquisition leading to an abundance of data 
\cite{Czimmermann2020sensors_survey,Yang2020defect_detection_survey,Wen2023steel_defdec_survey,Tang2023steel_products_survey,Chen2021defect_detection_survey}.
However, the large variety of production processes and object geometries impedes 
the development of one-fits-all solutions. Instead, dedicated data has to be 
collected for each specific use-case. A large variety of surface defect types 
can be observed that have to be used for the training of defect recognition 
models. Surface defects are often classified based on their look, shape, tactile 
feedback or material \cite{Czimmermann2020sensors_survey}.
However, the more safety-critical a defect is, the less frequently it is 
observed because it is avoided by all means in the manufacturing process. This 
introduces a strong imbalance in the dataset in the amount of non-defected and 
defected samples as well as between defect types.
Additionally, manual annotation and inspection planning are costly processes 
limiting the available annotated data and inducing label inconsistencies due to 
subjective decisions or just plain fatigue of the inspector.

Generation of synthetic image datasets offers an elegant way out since a large 
variety of events  can be simulated, together with a precise ground truth 
annotation.
Despite the wide range of synthetic data applications \cite{Nikolenko2019}, the 
tradeoffs have not yet been systematically evaluated. We observe two main 
research directions in synthetic data generation: \textbf{generative}, using AI 
models to produce synthetic data, and \textbf{rule-based}, using computer 
graphics simulation based on physics described by a well-defined set of rules. 
With a generative approach, we control a single realization, whereas with a rule-
based approach, we control the realization context, giving us far better 
control, reproducibility, and reliability. 

\begin{figure}[bt]
\centering
\includegraphics[width=.23\textwidth]{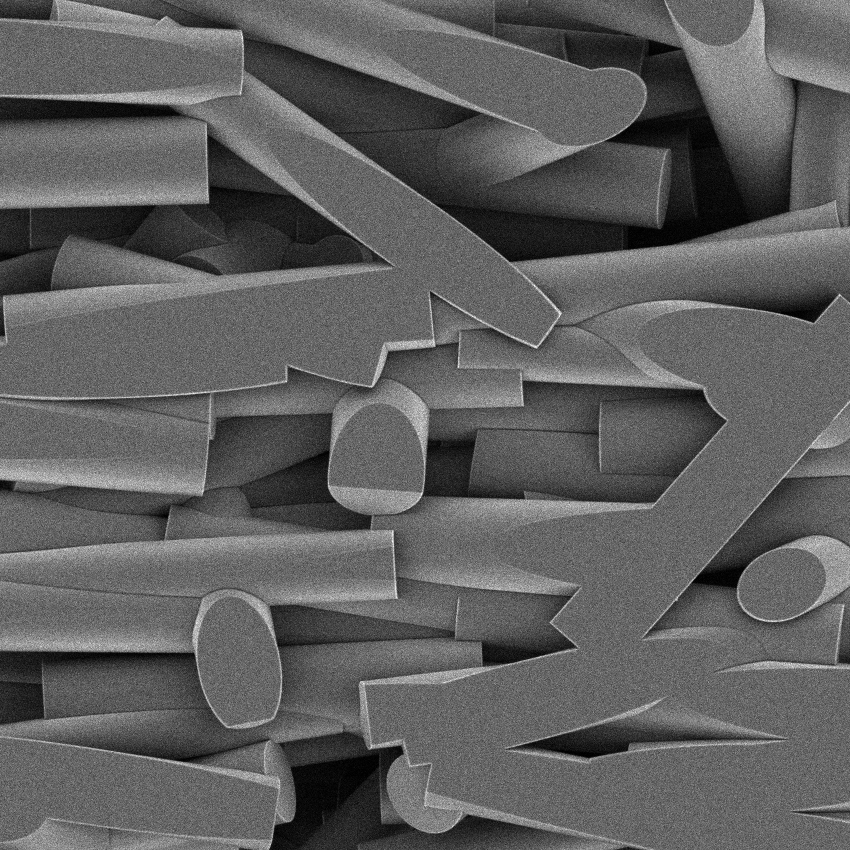}
\
\includegraphics[width=.28\textwidth]{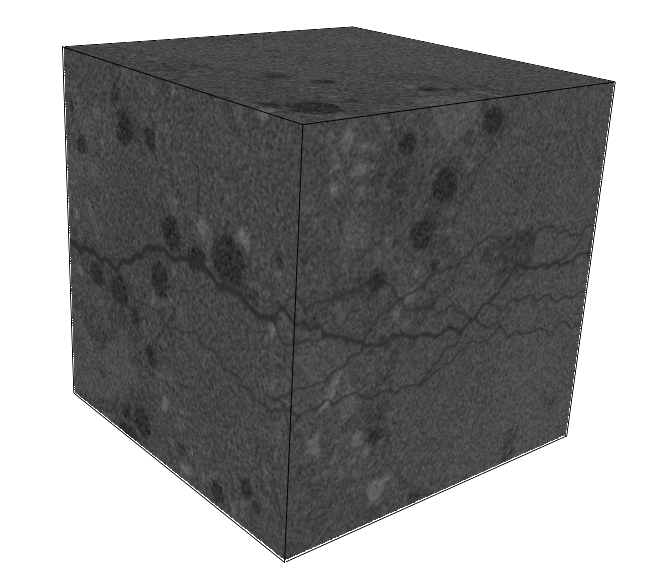}
\
\includegraphics[width=.28\textwidth]{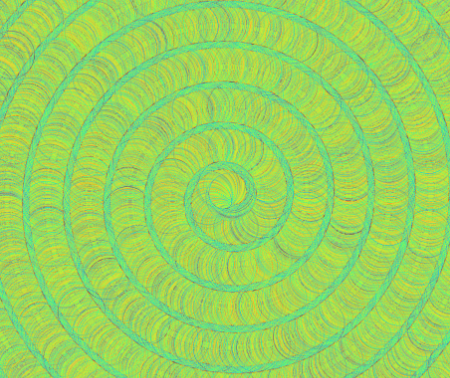}
\caption{Realizations of stochastic geometry models. From left to
  right: Simulated SEM image of a Boolean model of cylinders, 
  Sec.~\ref{sec:challenge-fib-sem}. CT image of a concrete sample with 
  impressed crack generated based on a Voronoi tessellation, 
  Sec.~\ref{sec:challenge-cracks}. Color coded height map of a 
  simulated milling pattern on a metal surface, Sec.~\ref{sec:challenge-vipoi}.}
\label{fig:examples-model-use}
\end{figure}

Generation of images using stochastic geometry models, see 
Figure~\ref{fig:examples-model-use}, is a rule-based approach, 
where the stochastic geometry model parameters are used to control the 
realization context. This enables generation of a wide variety of structure 
types. The within-structure variation is naturally captured by the 
stochastic nature of the models. The ground truth is obtained for free 
since we know exactly where and how structure is placed in the scene, making it 
almost effortless to produce the labels.
At the same time, many new questions arise. In particular, it is not clear at 
all, which characteristics of the real image data have to be met to 
which degree of fidelity. To address these questions at least partially, 
we describe our approaches to generate realistic optical images of 
metallic surfaces and computed tomography (CT) images of concrete with cracks. 
For the latter, ML methods are trained using dedicated crack structure 
models. Moreover, a new scale invariant neural network type based on Riesz 
transforms instead of convolutions drastically reduces the number of 
parameters to be trained and consequently the amount of training data needed.

\section{Challenge 1: Reconstruct highly porous structures from FIB-SEM images}
\label{sec:challenge-fib-sem}
SEM images feature an exceptional depth of view. For porous structures, this 
causes rather bright gray values originating from solid structure much deeper 
than the currently imaged plane. When stacking SEM images of consecutive 
slices cut by a FIB, these shine-through-effects turn into artifacts, 
rendering the decision which pixels really represent solid structure a very 
tedious one.

Synthetic images simulated based on very simple geometries, namely 
realizations of Boolean models of spheres or spherical cylinders, 
but correctly modeling major physical effects of the imaging method 
\cite{prill12:scanning} turned out to be very helpful in developing 
a morphological algorithm solving the task \cite{prill13:jmi} as 
well as quantitatively comparing methods 
\cite{salzer-prill-comp2015}. In terms of ML, reconstruction is a 
semantic segmentation task. 
Having the synthetic image data at hand, the question came 
naturally: Why not use them to train and test an ML solution?

Indeed, training the 3D version of the ready made, easily available
convolutional neural network 3d U-net \cite{3dunet}, yielded a usable 
solution \cite{fend2020}. Adaptions to the specific task consisted in 
particular in emphasizing the edges by a suitable weight map and a 
proper overlap of the patches. 
For learning how to segment real FIB-SEM images, the synthetic
training data nevertheless turned out to be too clean. This issue was 
solved by gray value histogram matching. However, it does not shed a 
good light on generalizability of the solution. 
A more systematic study on which features of the structures or the 
images have to be captured how well is currently under way.
\begin{figure}[ht]
\centering
\includegraphics[width=.18\textwidth]{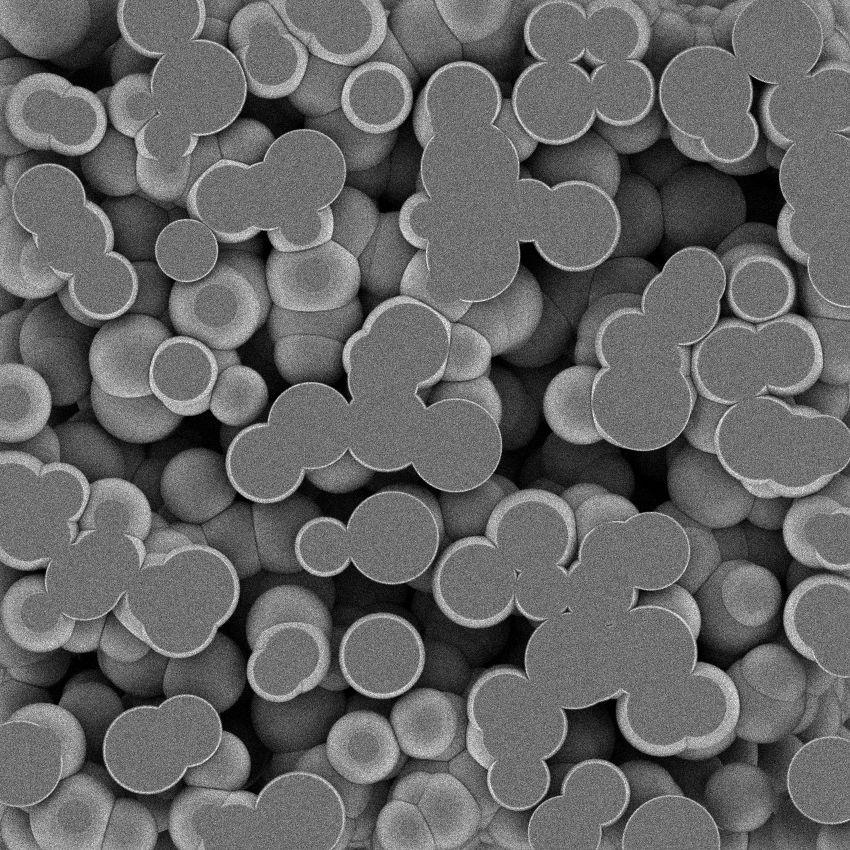}
\hfill
\includegraphics[width=.18\textwidth]{cylinders.png}
\hfill
\includegraphics[width=.18\textwidth]{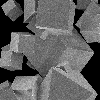}
\hfill
\includegraphics[width=.18\textwidth]{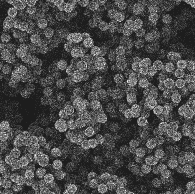}
\hfill
\includegraphics[width=.18\textwidth]{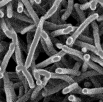}
\caption{Synthetic SEM images of realizations of Boolean models of
  spheres, cylinders, and cubes, a Cox-Boolean sphere model, and 
  Altendorf-Jeulin's curved fiber model \cite{altendorf10:model}. 
  Images generated using Prill's method \cite{prill12:scanning}.}
\label{fig:synthetic-sem}       
\end{figure}

\section{Challenge 2: Detect and segment cracks in 3D images of concrete}
\label{sec:challenge-cracks}
Semi-synthetic image data is the only option to get 3D images with cracks 
accompanied by a consistent ground truth. Generating 3D training data 
by interactive annotation is not really feasible as cracks in CT images 
can be very thin and appear rather dark without a strong contrast to
the surrounding material. The concrete matrix is heterogeneous, the gray value distribution 
there may therefore vary strongly.
Completely synthetic images are far from the real ones as the simulation 
tools cannot capture this heterogeneity properly.

Two types of synthetic cracks have proven to work, realizations of
fractional Brownian surfaces \cite{brownian-surface-matlab} and minimal surfaces formed by facets of a Voronoi tessellation 
\cite{jung22cracks}. The thickness can be nicely controlled by adaptive 
dilation. Synthetic cracks blend into the real background by
distributing gray values as observed in air pores and 
smoothing the edges to mimic the partial volume effect.
\begin{figure}[t]
\centering
\includegraphics[width=.24\textwidth]{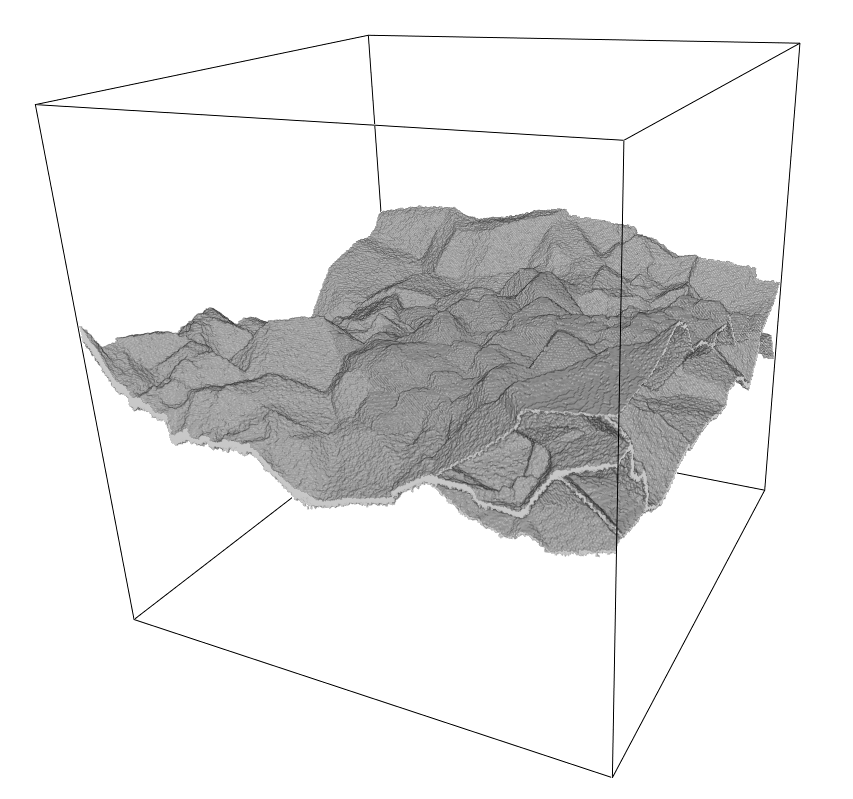}
\hfill
\includegraphics[width=.24\textwidth]{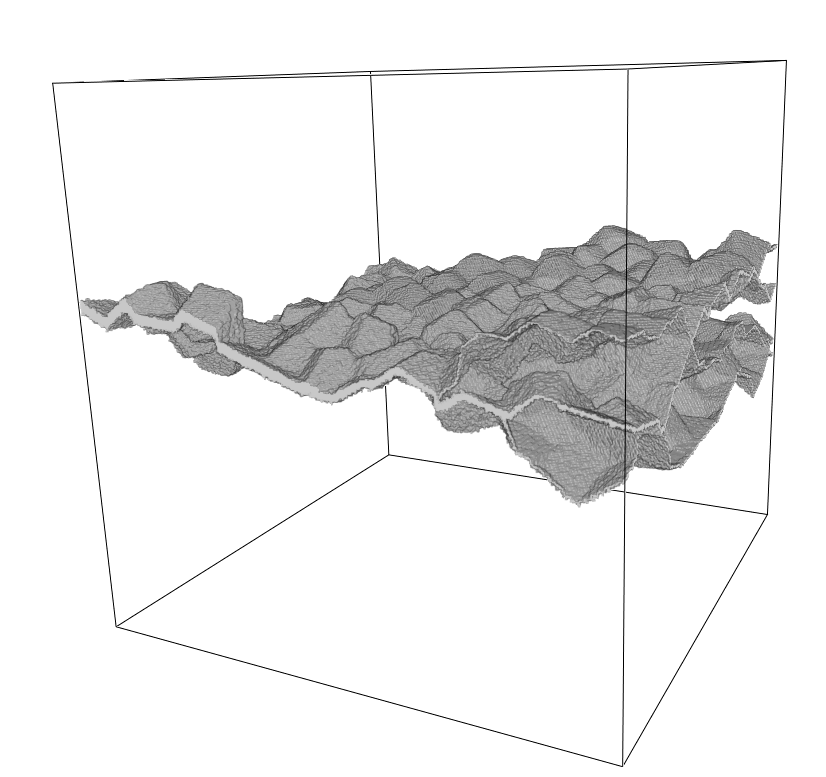}
\hfill
\includegraphics[width=.24\textwidth]{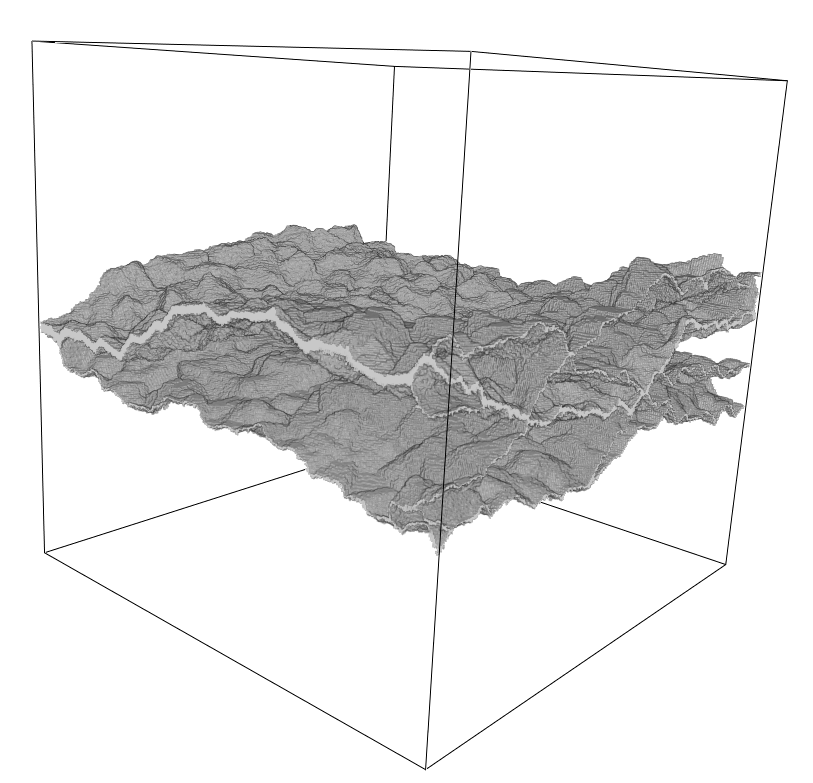}
\hfill
\includegraphics[width=.24\textwidth]{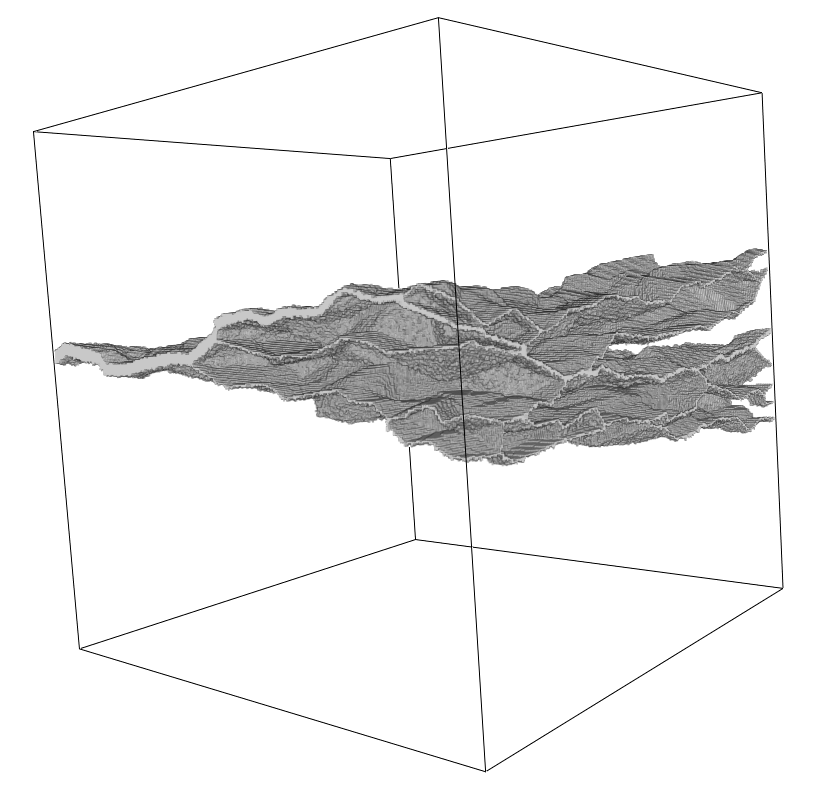}\\
\includegraphics[width=.24\textwidth]{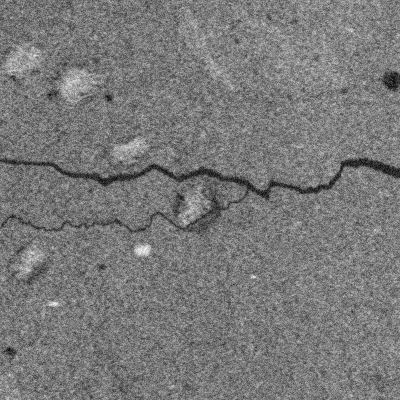}
\hfill
\includegraphics[width=.24\textwidth]{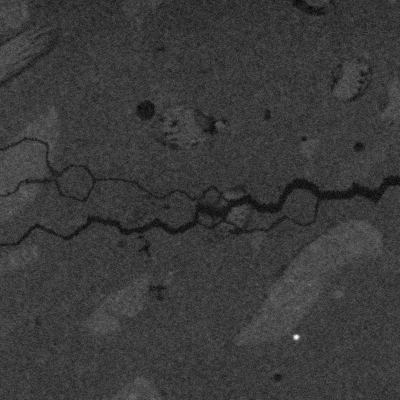}
\hfill
\includegraphics[width=.24\textwidth]{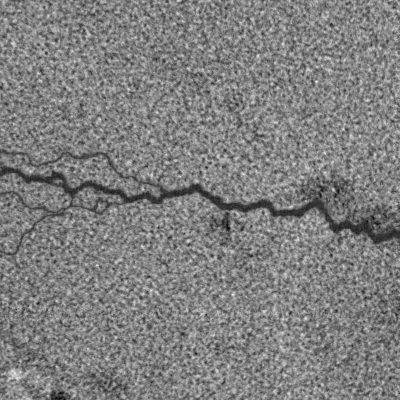}
\hfill
\includegraphics[width=.24\textwidth]{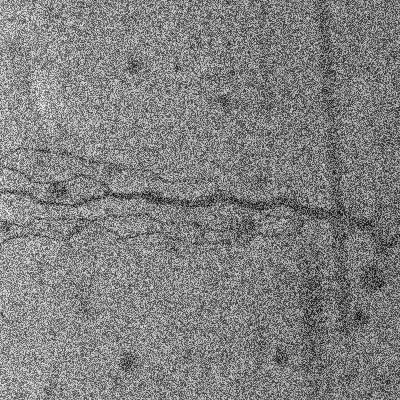}
\caption{Examples of simulated cracks in $400^3$ voxel images. Minimal
  surfaces from spatial Voronoi tessellations generated by a
  Mat{\'e}rn cluster process (left), force-biased sphere packing
  (center left), a Poisson point process (center right), and a 
  Poisson point process stretched in x and z directions (right). 
  The crack widths vary according to a Bernoulli 
  random walk~\cite{jung22cracks} with parameter $p=0.01$. 
  Top: volume renderings. 
  Bottom: 2D slices of the 3D images of the cracks superimposed on CT
  images of normal, high performance, and aerated concrete and the 
  polypropylene fiber reinforced concrete from \cite{ict23}. 
  Voxel sizes are $23\,$\textmu m, $20\,$\textmu m, $2.8\,$\textmu m, 
  and $60\,$\textmu m. Hence, cube and slice edge lengths vary between $1$ and $24\,$mm.
}
\label{fig:crack-examples-synthetic}
\end{figure}

So far, there is no provable advantage of ML over classical methods for 
crack segmentation. Hessian-based percolation \cite{paetsch11, paetsch12, 
barisin22} and template matching \cite{paetsch19} still have their 
advantages \cite{ict23}. However, from the application 
point of view, fine-tuning a trained ML model to deal with a yet unseen 
concrete type is probably easier than optimizing the parameters of the 
classical methods.

\begin{figure}[t]
\subfloat[original]{\includegraphics[width=0.24\textwidth]{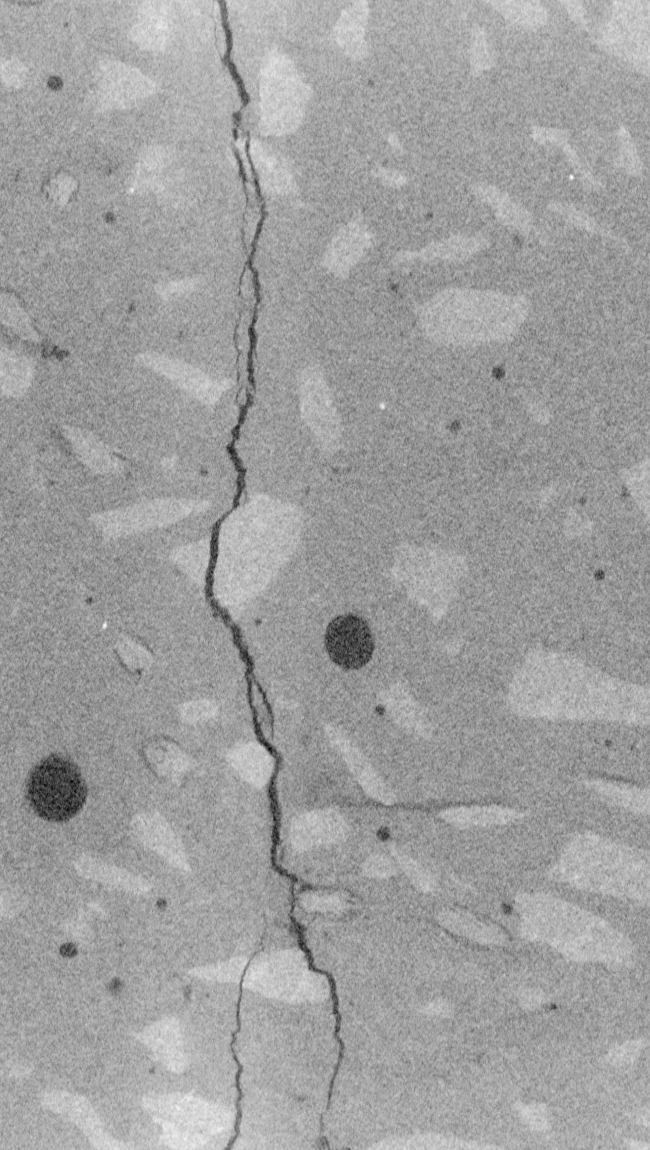}}
\hfill
\subfloat[Hessian-based percolation]{\includegraphics[width=0.24\textwidth]{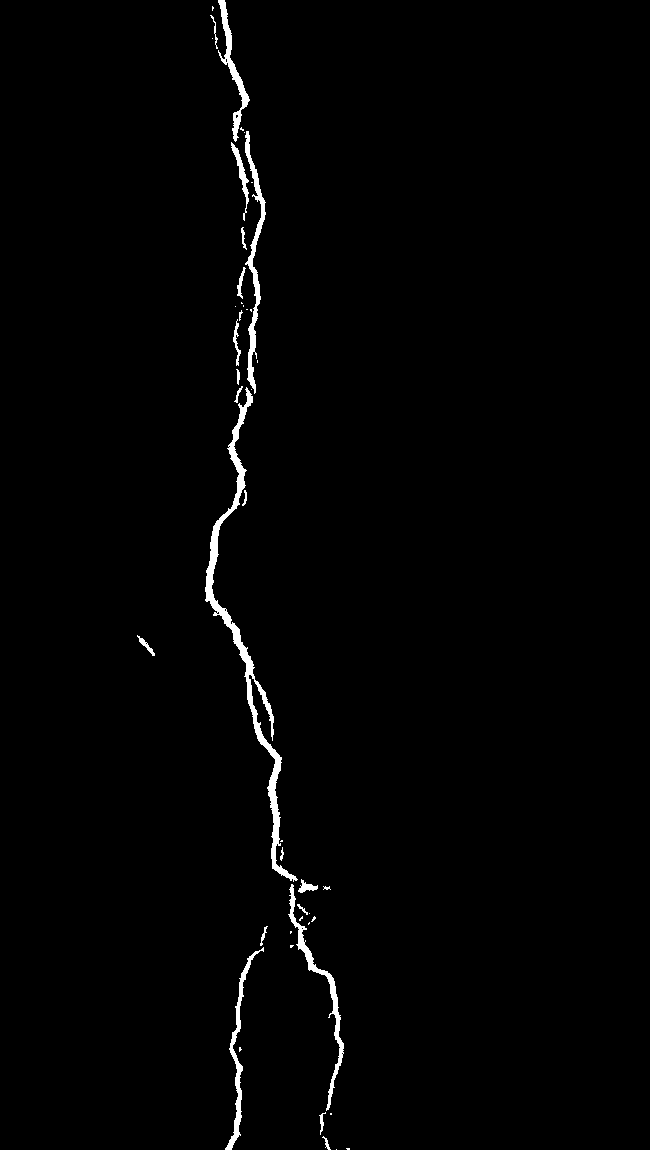}}
\hfill
\subfloat[3d U-net]{\includegraphics[width=0.24\textwidth]{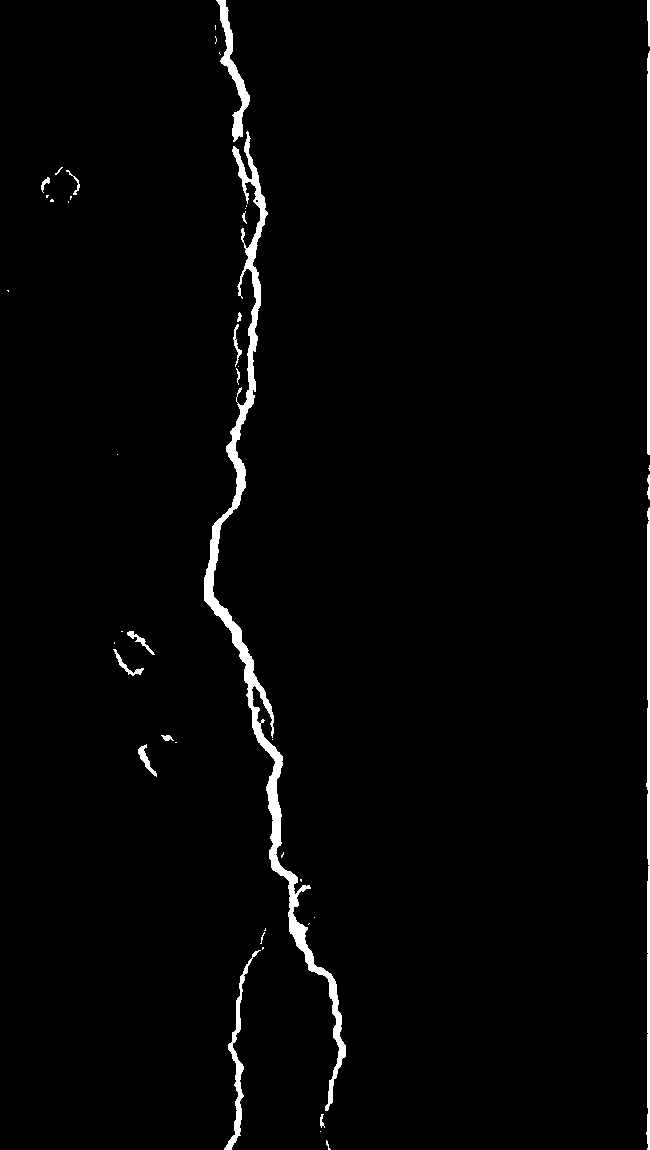}}
\hfill
\subfloat[RieszNet, median filtered]{\includegraphics[width=0.24\textwidth]{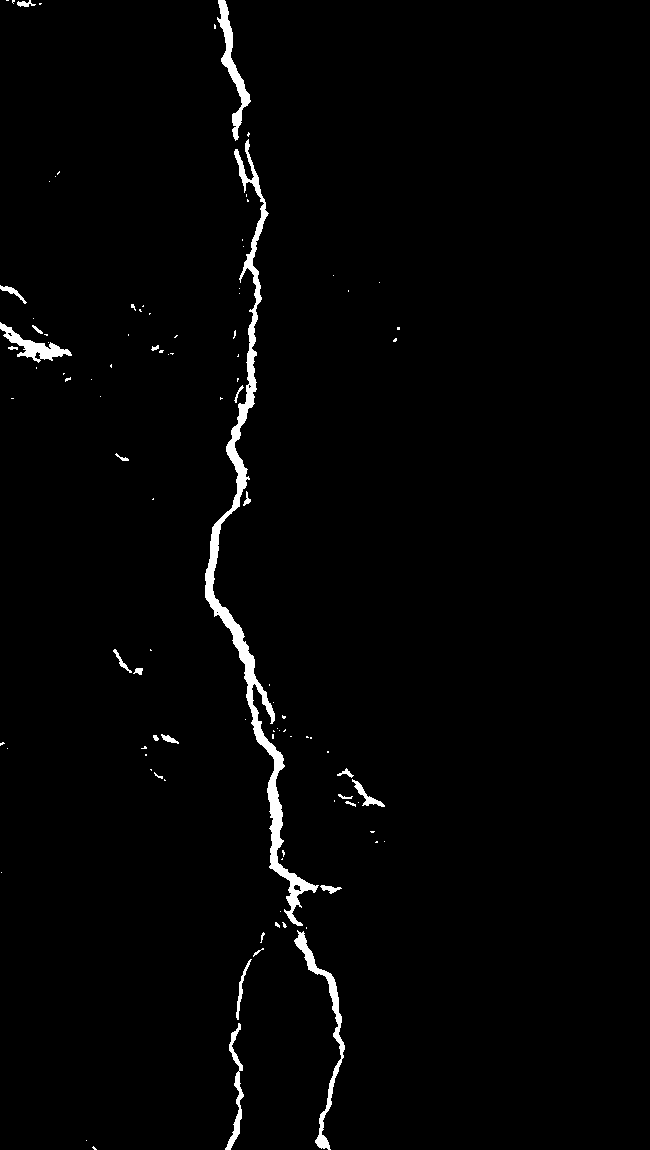}}\
\caption{Segmentation results for the high performance concrete sample from \cite{ict22}.}
\label{fig:results-nc}      
\end{figure}

\begin{figure}[t]
\subfloat[original]{\includegraphics[width=0.24\textwidth]{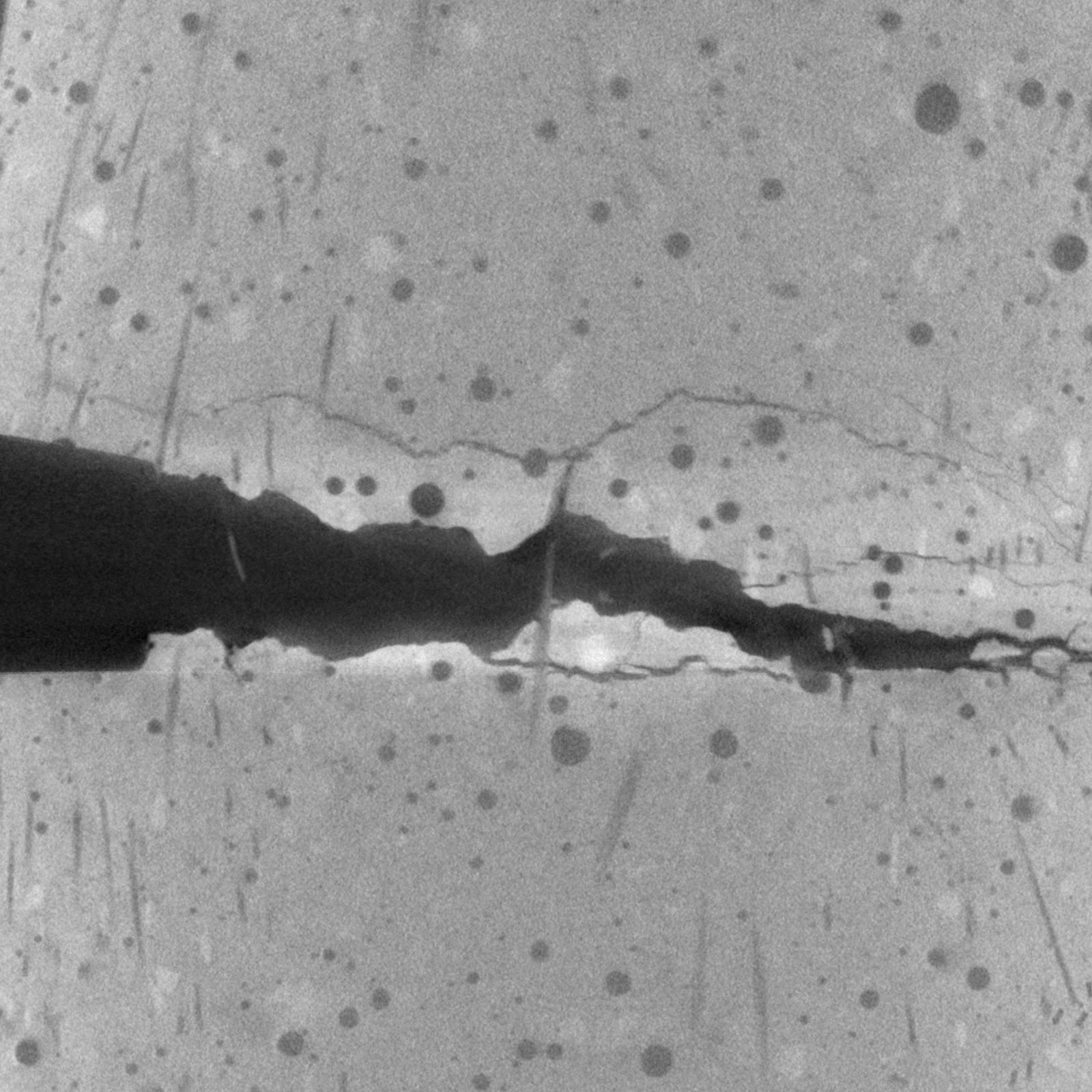}}
\hfill
\subfloat[Hessian-based percolation]{\includegraphics[width=0.24\textwidth]{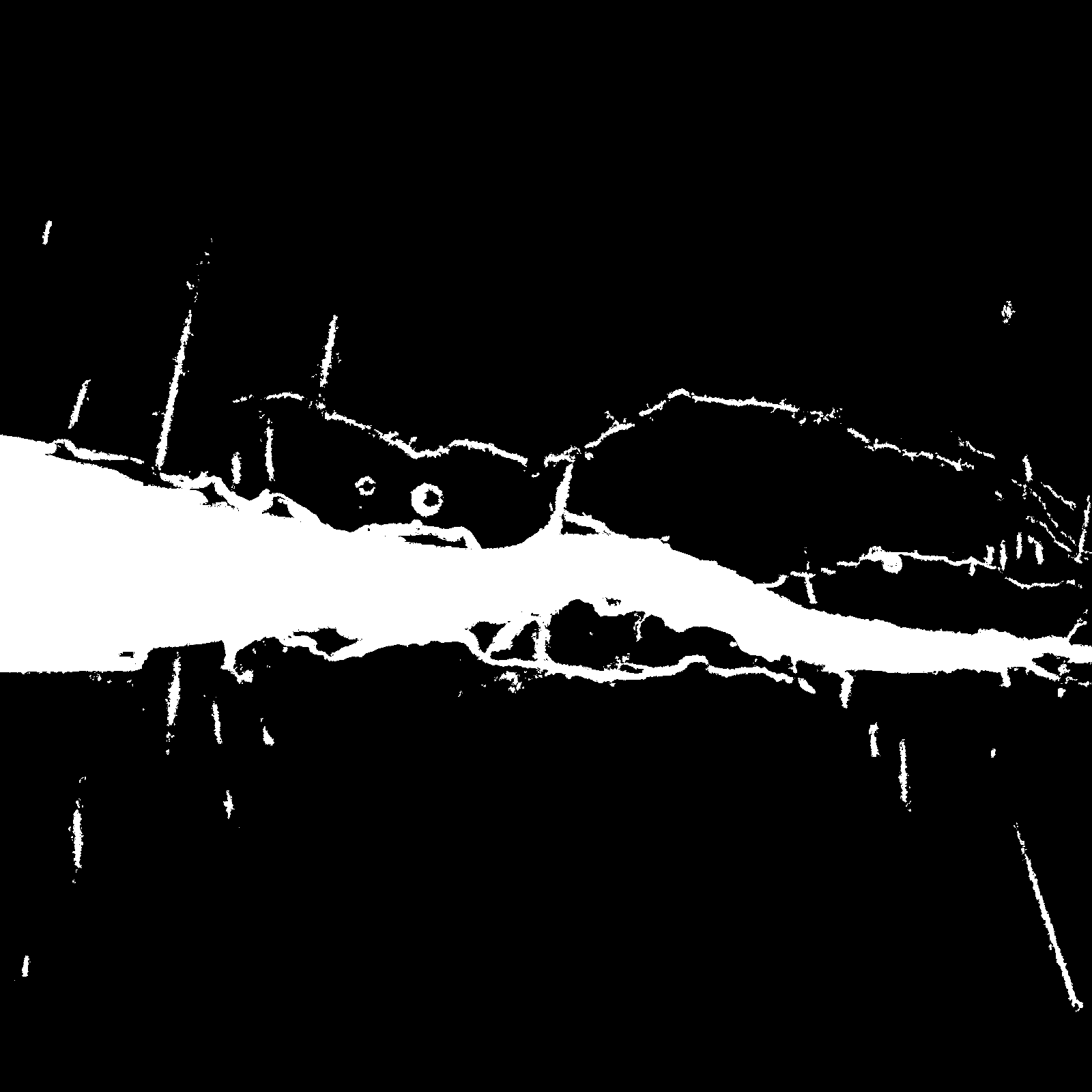}}
\hfill
\subfloat[3d U-net]{\includegraphics[width=0.24\textwidth]{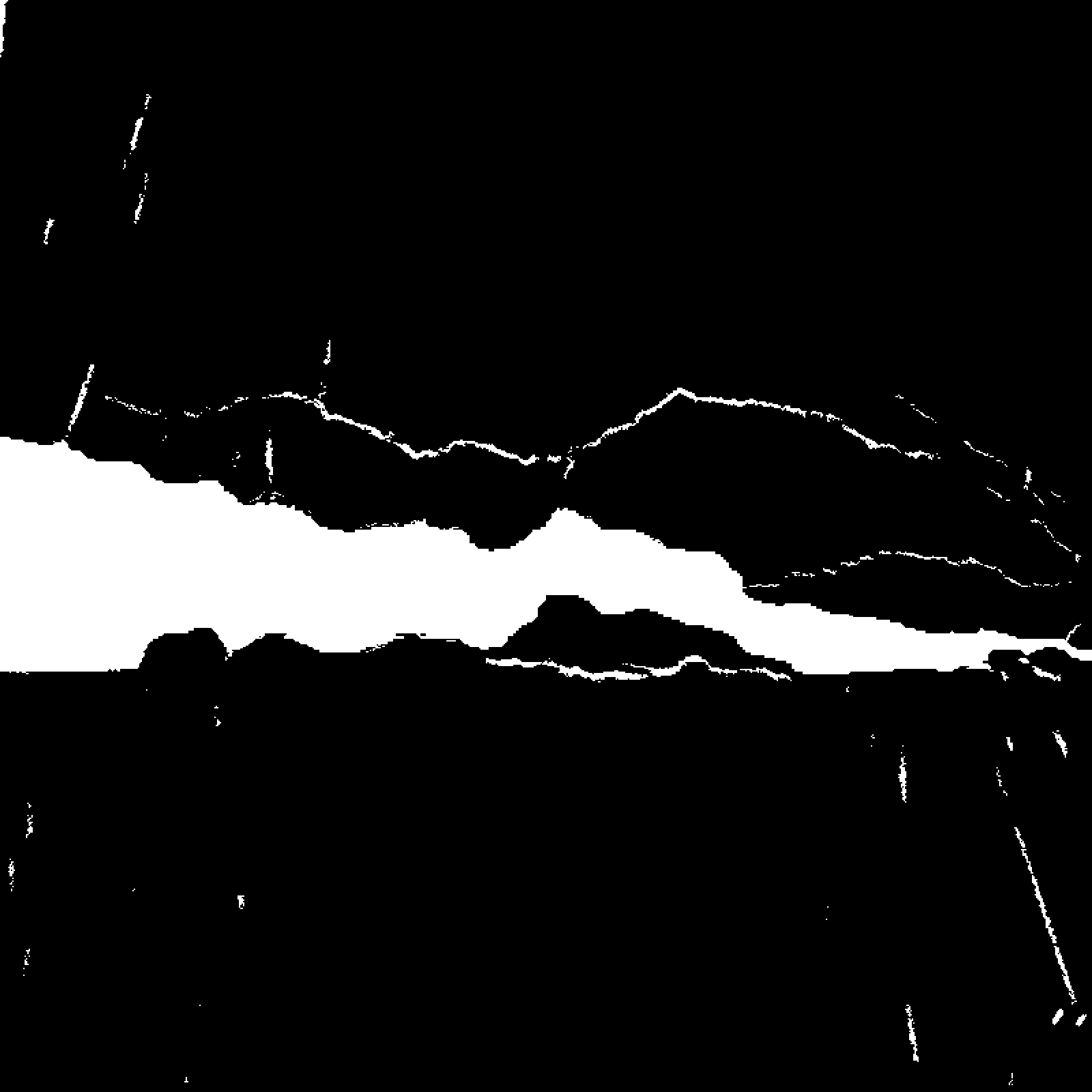}}
\hfill
\subfloat[RieszNet]{\includegraphics[width=0.24\textwidth]{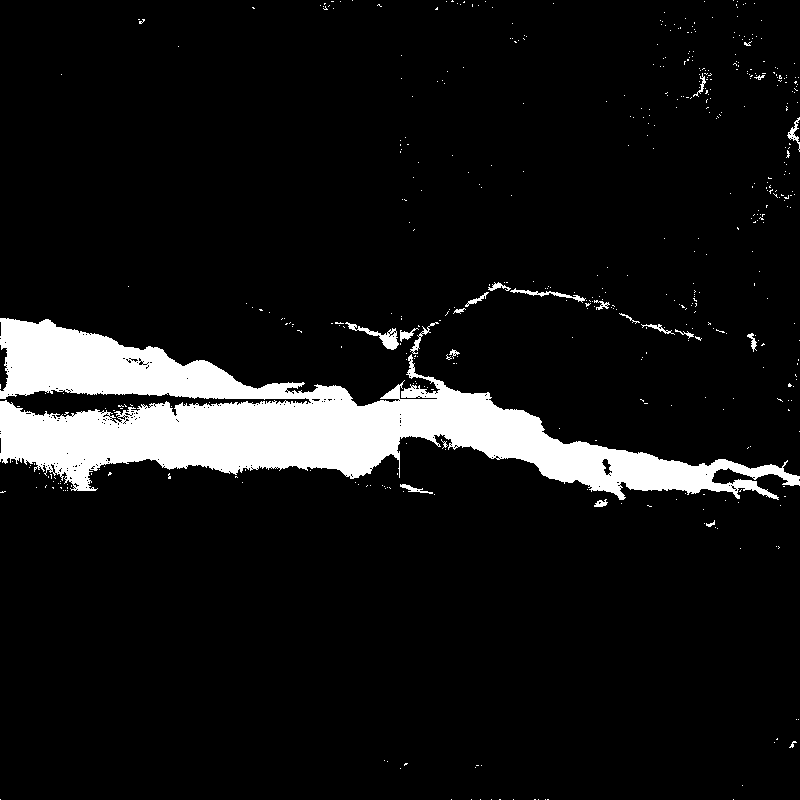}}
\caption{Segmentation results for the polypropylene fiber reinforced
  concrete sample from \cite{ict23}. The RieszNet result is obtained
  on a downscaled version of the image and stitched. The missing part 
  in the center of the crack is due to edge effects and the stitching.}
\label{fig:results-ppfrc}      
\end{figure}

The RieszNet \cite{barisin22riesz} is a completely new network type whose 
development was motivated by the crack segmentation task and 
the multi-scale nature of real cracks which can easily span a thickness range from 1 
to 200 voxel edge lengths \cite{ict23}. 
These multi-scale cracks can be taken care of by downscaling the image, 
applying the algorithm on each scale, and combining the results by 
upscaling again \cite{ndt-ce,ict22}, or by training additionally on 
synthetic multiscale cracks generated by adaptive dilation 
\cite{nowacka23}, see Figures~\ref{fig:results-nc} and 
\ref{fig:results-ppfrc} for exemplary results. 
Both strategies are costly and have obvious limitations in the thickness 
range that can be covered. RieszNet instead replaces the convolutions in a 
typical convolutional neural network like 3d U-net by 1st and 2nd order 
Riesz transforms.
The scale invariance of the latter two carries over to the network 
\cite{barisin22riesz}. This results not only in a significant reduction in 
computation time, it also drastically reduces the number of trainable 
parameters by two magnitudes and thus allows for much smaller training 
data sets. The scale invariance of the RieszNet is not just a theoretical 
promise but proves to work in practice. The exemplary RieszNet results 
shown in Figures~\ref{fig:results-nc} and \ref{fig:results-ppfrc} are achieved by 
training solely on synthetic cracks of fixed width 3 voxel edge lengths.

To summarize, well-known CNN, properly trained and fine-tuned do the job, 
are however costly if image scaling is needed and require fine-tuning for 
every new concrete type. The strong potential of Barisin‘s RieszNet has to 
be explored more deeply and better used by an implementation of the
Riesz transform allowing to process full-scaled 3D images and dedicated post-processing.

\section{Challenge 3: Virtually plan and simulate optical surface inspection}
\label{sec:challenge-vipoi}
Optical surface inspection tackles a large variety of defect types,
depending greatly on the manufacturing process. Several research
studies investigating surface defects from different perspectives give
an impression of the variability
\cite{Czimmermann2020sensors_survey, Erdogan2015, Tang2023steel_products_survey}.

A major challenge in detecting surface defects is 
the strong variation in the frequencies at which certain defect types
occur in production. 
This can cause severe imbalances in datasets acquired in real industrial settings.
The imbalance can be detected by manual data analysis, only.
Such analysis is however quite impractical since the image count is
ideally in thousands and it would require the evaluator to imagine all 
potentially possible defect scenarios. 
This complicates inspection model development and causes bias and lack
of model robustness for defects which have not been observed sufficiently often.

One particularly demanding use case is visual inspection of specular
metal surfaces. Defects can significantly change appearance depending
on imaging conditions like light and camera placement and parameters. 
This increases the number of detection cases that need to be handled even further.
Thus, deep learning models are needed that offer higher recognition
robustness in such complex cases. The robustness can only be achieved
by representing a wide variety of defect appearances for each defect class.
As explained earlier, collecting such a dataset is highly impractical, 
introducing the need for synthetic generation of industrial data.

Generative models have been used for creating surface inspection data 
\cite{Wang2022dtgan,Wen2023steel_defdec_survey}. 
However, hallucination is an inherent behavior which can not be
reliably controlled. Schemedemann \cite{Schemedemann2023}  combined
rule-based generation with the generative approach to increase the
realism of the images and reported both hallucination and possible
dataset degradation. It can happen that the image looses the defect
after the generative domain adaption, although the defect label is
still present in the ground truth. Or the generative model creates a
defect where the ground truth does not account for the defect. There
is no automatic way to catch these artifacts. Hence, a human operator 
would have to be employed once again to manually check the images.

Procedural computer graphic models on the other hand enable explicit 
control over the geometry and microstructure of defects and surface
finishes \cite{bosnar2022texsynth,Moonen2023cad2render} and thus 
rule-based synthetic data creation.
Procedural methods enable the design of functions which can
realistically model defect geometry \cite{bosnar2023defect_modelling} 
and micro-structures of surface texture \cite{bosnar2022texsynth}. 
The models themselves can be created using any mathematical modeling
approach, and image rendering techniques can use these methods to
generate realistic images, with variability controlled by fitting
parameters to physical measurements (e.g. elevation and density of the
surface pattern). This has been introduced and well documented in the
pipeline introduced by 
Bosnar et al. \cite{Bosnar2020pipeline}.

\begin{figure}[t]
    \centering
    \subfloat[Spiral milling]{
        \centering
        \includegraphics[width=0.23\textwidth]{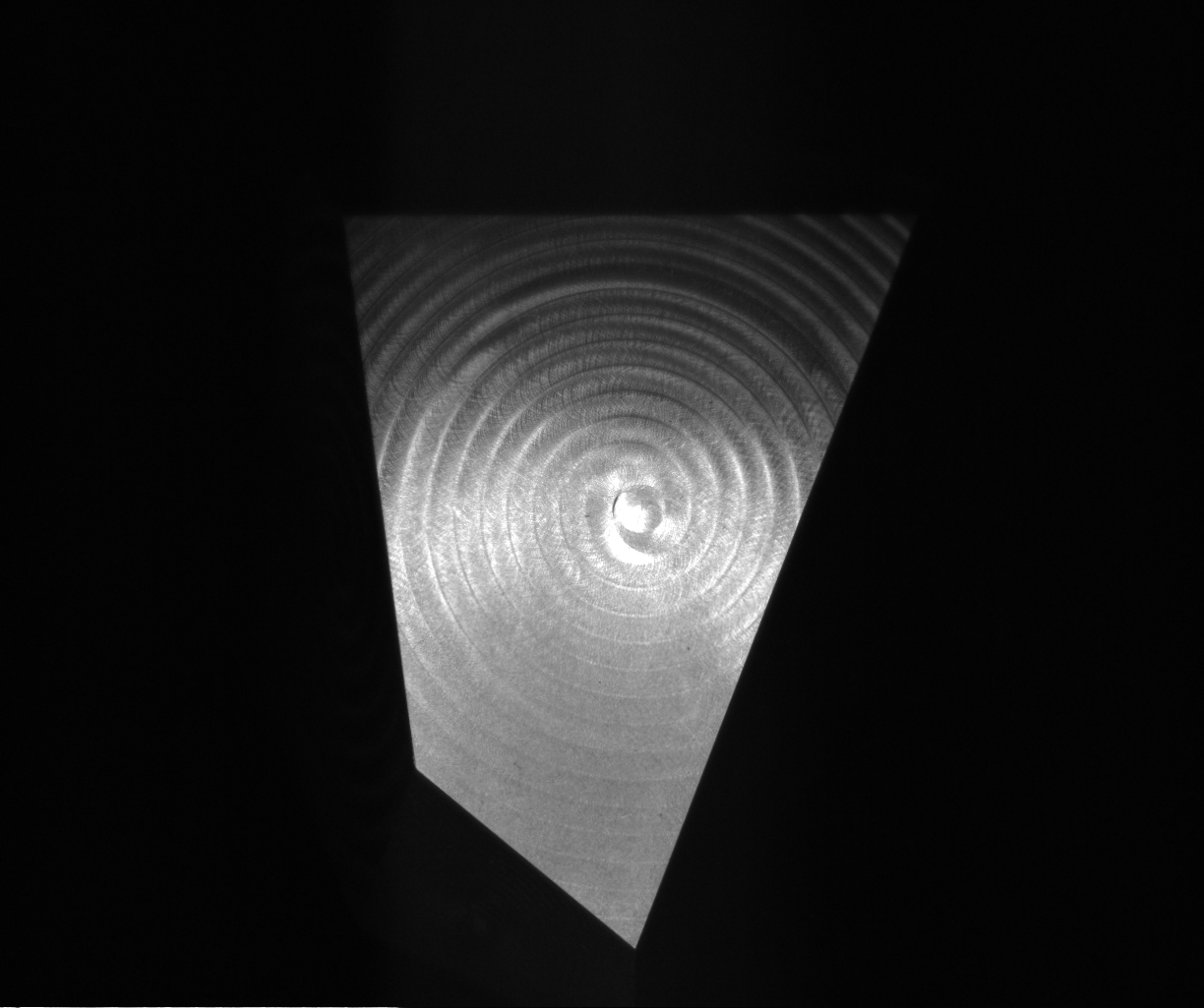}
        \hfill
        \includegraphics[width=0.23\textwidth]{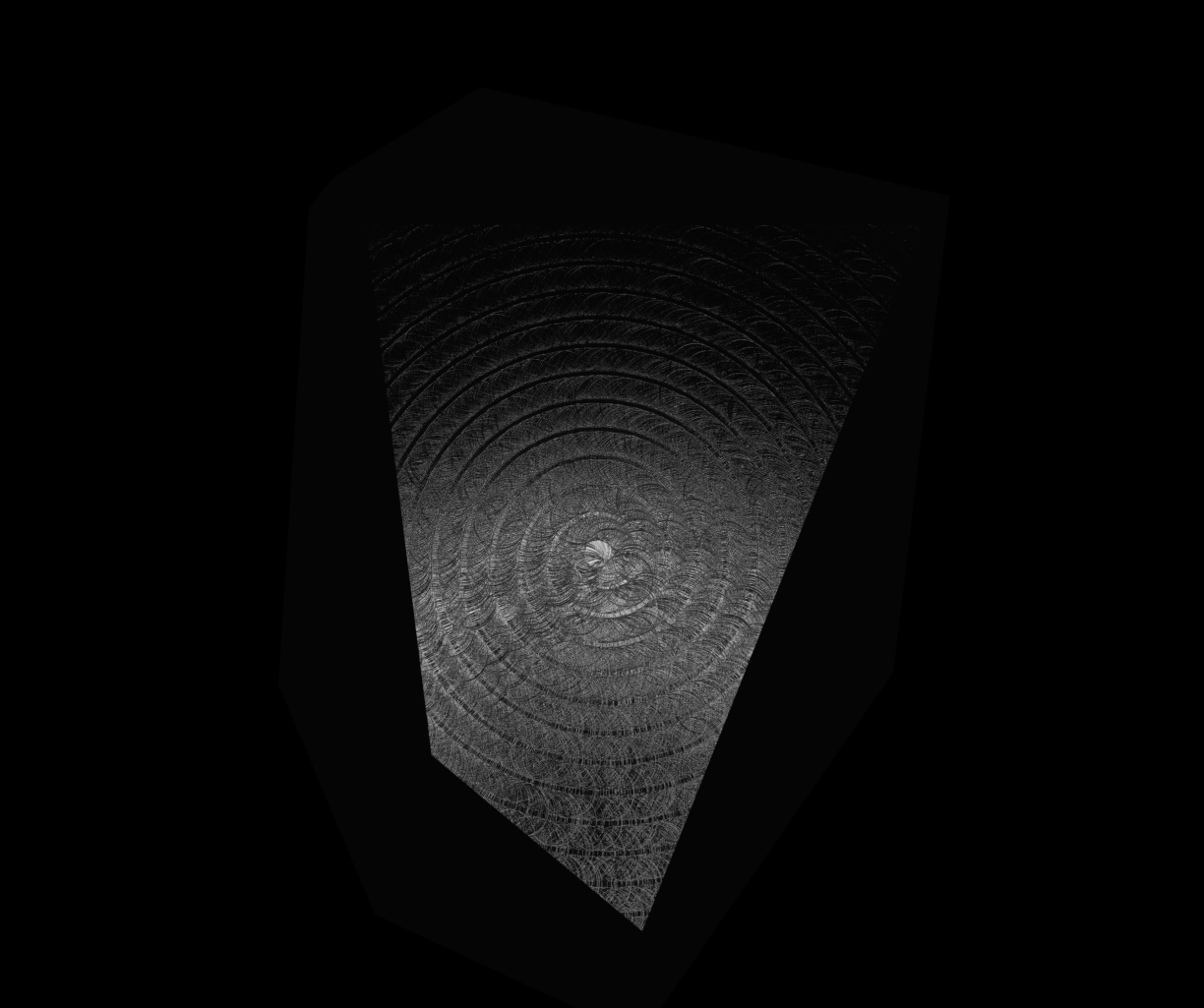}
        \hfill
        \includegraphics[width=0.23\textwidth]{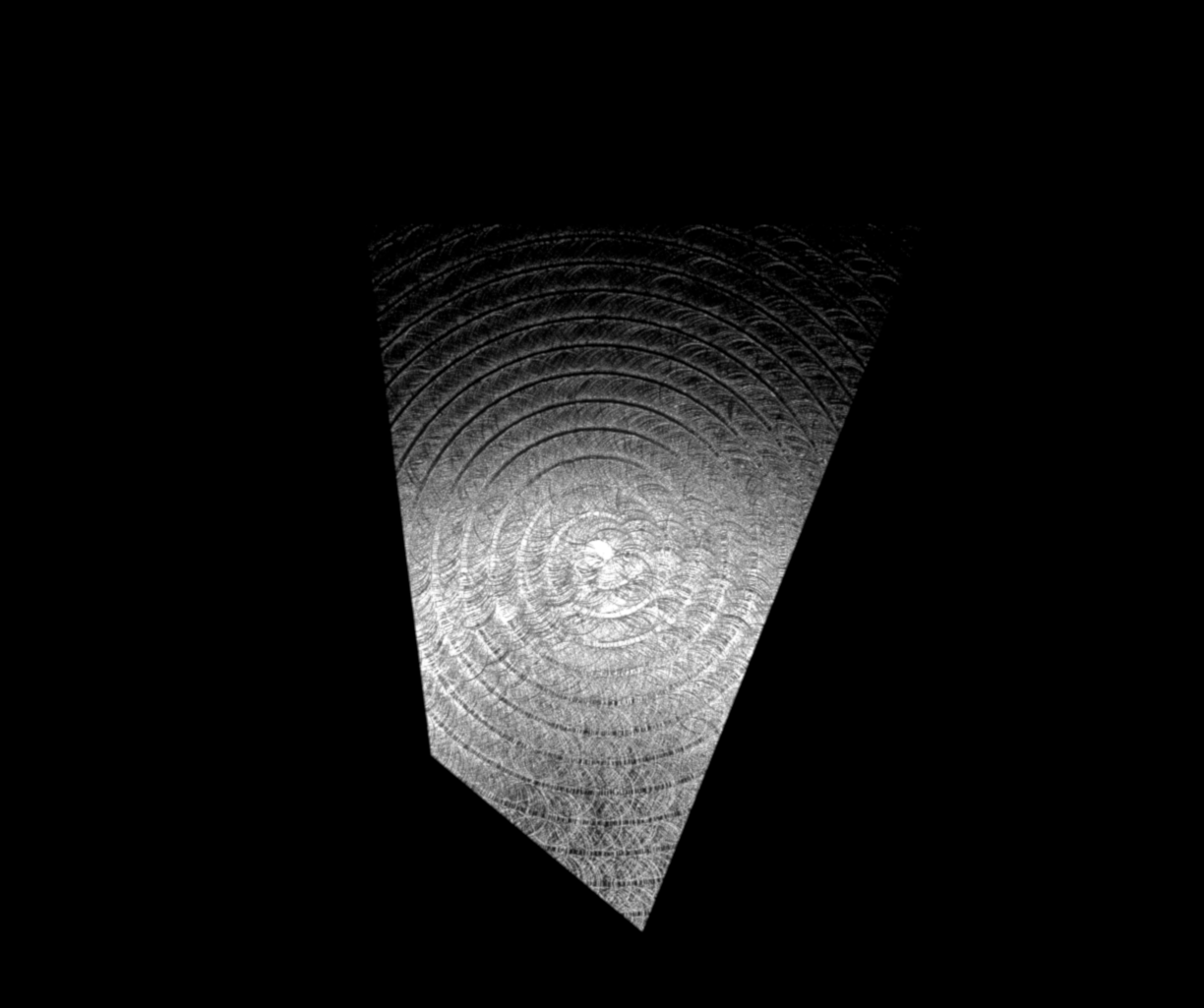}
        \hfill
        \includegraphics[width=0.23\textwidth]{spiral_innen_crop.png}
        \hfill
        \includegraphics[width=0.04\textwidth]{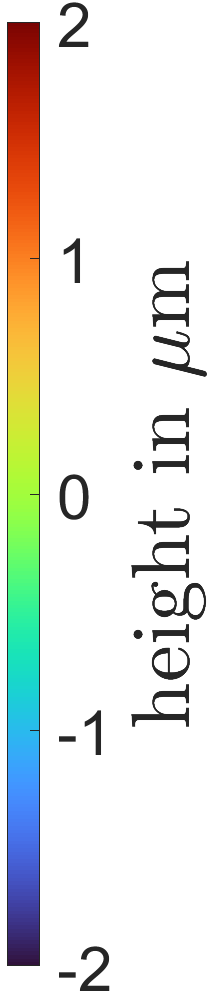}}\\
    \subfloat[Parallel milling]{
        \centering
        \includegraphics[width=0.23\textwidth]{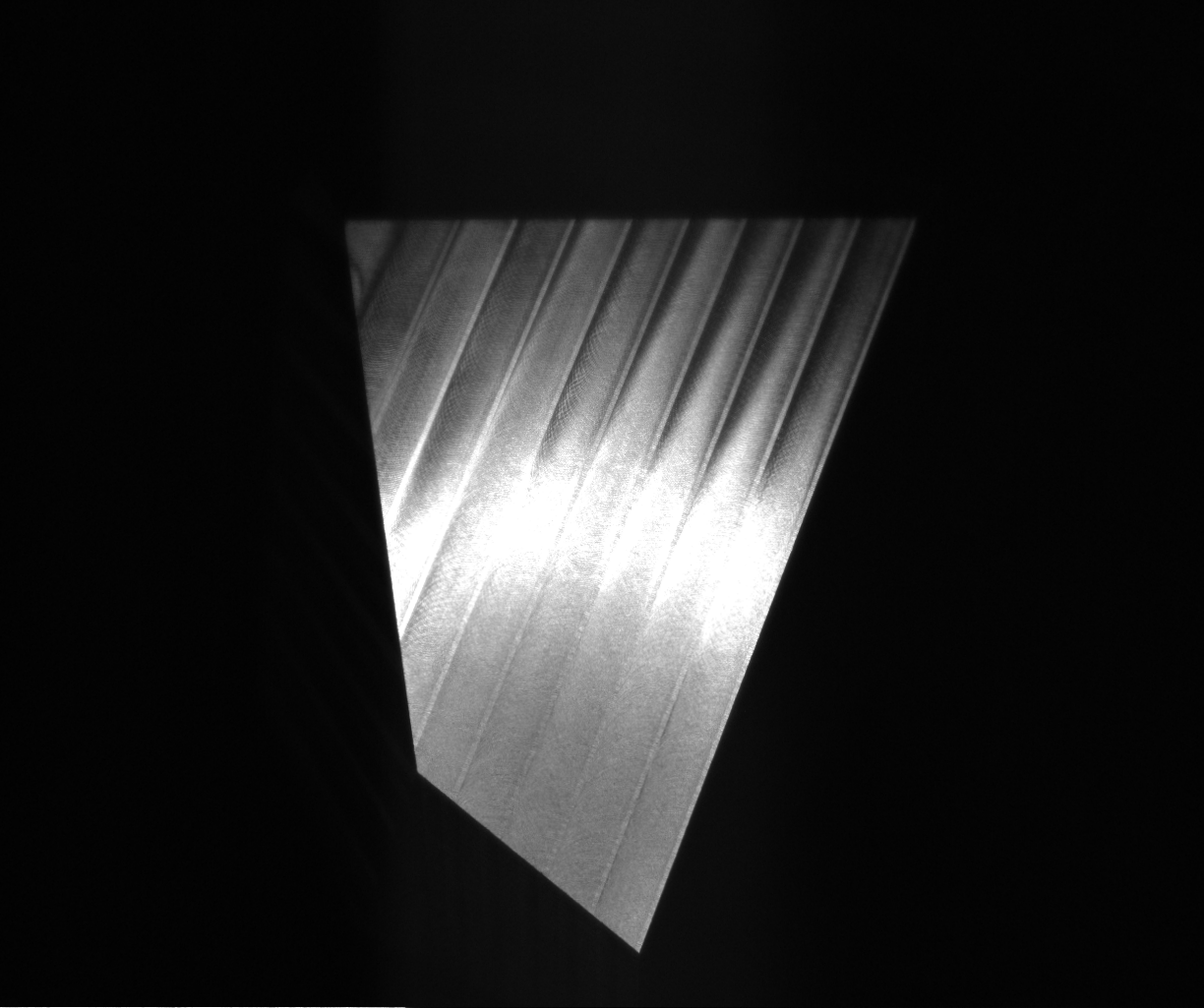}
        \hfill
        \includegraphics[width=0.23\textwidth]{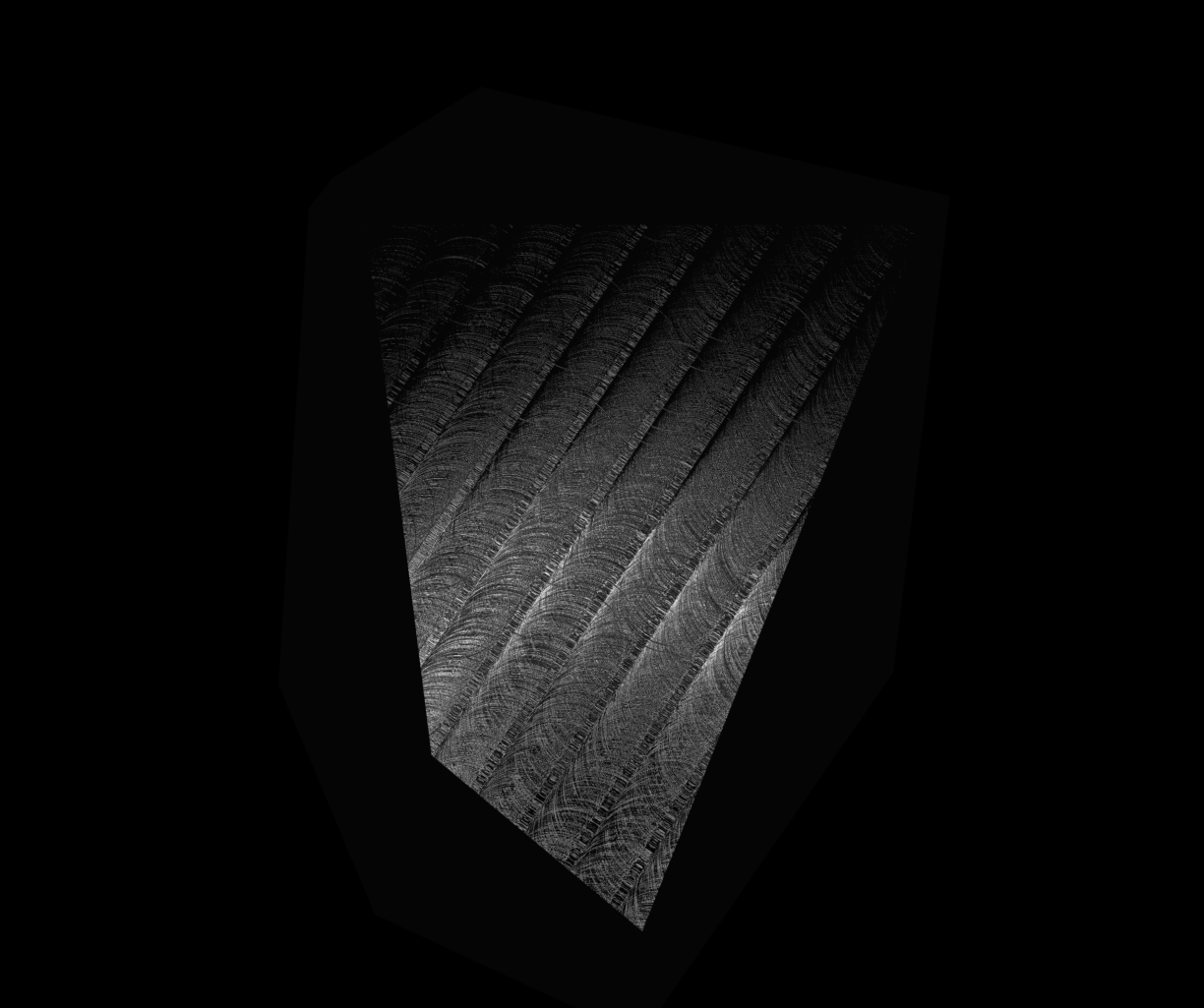}
        \hfill
        \includegraphics[width=0.23\textwidth]{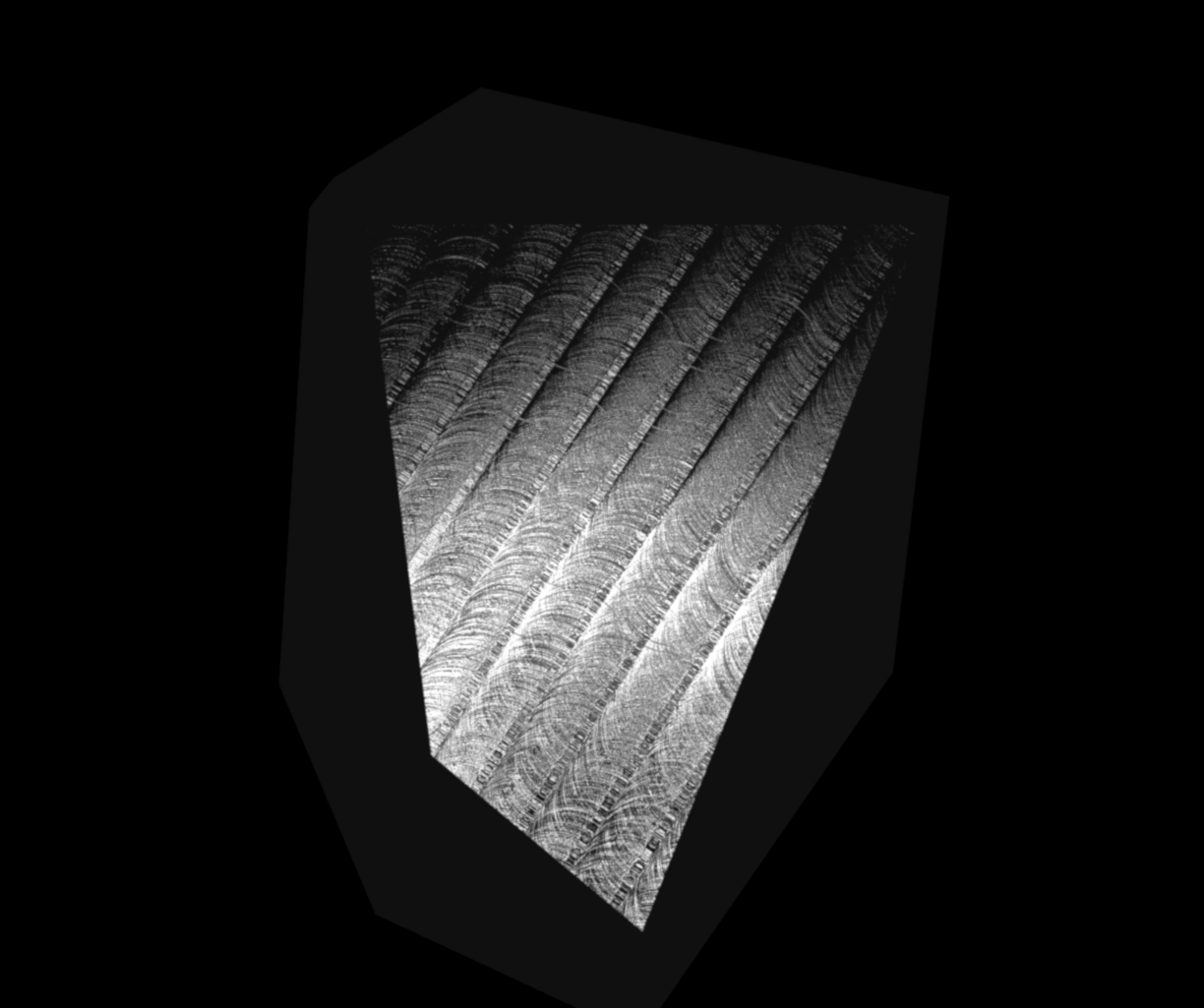}
        \hfill
        \includegraphics[width=0.23\textwidth]{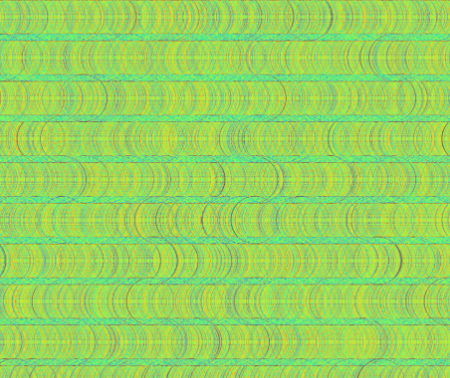}
        \hfill
        \includegraphics[width=0.04\textwidth]{colorbar2.png}}
    \caption{Comparison of real image, synthetic image, and synthetic
      image with additional exposure adjustment for two milling
      paths. The rightmost images feature the procedural textures used
      to generate the synthetic data, cropped to make the texture
      structures more visible.}
    \label{fig:synosis_real_sim}      
\end{figure}

So far, the use of stochastic geometry models has shown to be very promising for achieving a high degree of realism. 
The work of SynosIs \cite{SynosIs_in_preparation} aims to create synthetic images of machined aluminum surfaces. 
There the focus lies on milling which is a common method to form objects.
Material is removed by blades installed at the milling head while the head rotates.
Meanwhile, the tool moves along a predetermined path over the surface.
On the final surface, the milling creates a pattern of micrometer-scale surface scratches.
This pattern as well as how it interacts with optical light depend on
size, path, and speed of the cutting tool. 

Additionally, due to material inhomogeneities, cutting tool wear, and
environmental influence, no two surfaces are completely the same, even though they might be produced using exactly the same parameters.
To recreate this behavior, stochastic geometry models are integrated into Bosnar's pipeline for modeling both the defects and the textures procedurally. 
The defects are modeled as described in \cite{bosnar2023defect_modelling}. 

The texture model is designed to mimic the physical change of the surface due to milling.
The scratches defining a face-milled pattern are ring-shaped caused by the cycloidal paths of the blades.
To simplify that movement in the model, rings are used instead.
A multi-step model is developed describing the positioning of rings, the appearance of an individual ring, and the interaction of overlapping rings.
Moreover, the model is parameterized including the knowledge of real-world parameters from the manufacturing process.
The lateral cutting depth, the feed rate, and the spindle speed determine the tool-path and are therefore decisive for the placement of rings.
The diameter and tilting of the milling head and the width of its blades directly describe the fundamental shape of the rings.
All parameters not known from the manufacturing process are estimated
based on visually comparing the 
synthetic textures and real height measurements of the milled surfaces.
The observed variation within textures generated using the same parameter configuration is captured selecting their values from suitable random distributions.

Fitting the parameter values to the machining parameters and to meet
microscopic images of the surfaces, it is possible to synthesize an
image in which the object's surface texture is causing a very similar
light response and pattern when compared with the real image (see
Figure \ref{fig:synosis_real_sim}). Note that the illumination
intensity for these images has been matched approximately. By
increasing the exposure manually, the similarity of overexposed
regions becomes more prominent, rendering the visual appearance even
more similar.

\section{Conclusion and discussion}

Synthetic data has proven to benefit the training of machine learning
models for segmentation and classification problems
\cite{fend2020,fulir2023comparisson_kupplung,Moonen2023cad2render,nowacka23}. Here,
we emphasized in particular the advantages of 
building the synthetic data on stochastic geometry models. There are 
nevertheless new challenges arising. In general, we still do not know
what degree of realism has to be reached, neither in terms of
structure, texture, or pattern nor in terms of the overall synthetic
image appearance and acquisition-specific image features such as noise pattern or the shine-through effects in SEM.
It is however also a question, how much effort it takes to reach a
certain degree of realism, whether it is always necessary, and whether we can afford it.

It is possible to visually asses the synthetic image data and to state
that it is similar to the real data. However, there are so far no
general similarity metrics quantifying the deviations of synthetic and
real images for our challenges. Commonly used image
comparison metrics typically focus on super-pixel level features. 
Even when they do consider high frequency texture details as well
(e.g. SSIM), they do not have the capacity to correlate them to the global image pattern.  
This means for example that the metric would show very little
difference between the parallel and spiral milling patterns in 
Challenge 3 because those two textures are locally very similar. 

By developing dedicated metrics, it will be possible to evaluate
synthetic dataset quality beyond the empirical measure "the network
has successfully managed to generalize from it". This will open the
opportunity to quantify the so called domain gap -- 
something that has not been achieved so far. Domain gap refers to
differences in data characteristics and distribution between source and target domains \cite{Ben-David2006}. 
In the general context of machine learning, it is typically the
difference between the training dataset and the real data the 
model should be used on. In a more specific context of synthetic data it describes how realistic the synthetic data is.

In surface inspection, a particular challenge beyond realism measurement is over-
labeling. This term describes the situation where the ground truth reports defects 
that are unrecognizable for human inspectors. For instance, defects might be 
located in very dark areas or hidden by heavy blur, such that they should be 
detected with low confidence or not at all. However, defining a suitable defect 
visibility metric is still an open problem. Hence, the over-labeling problem has to be handled specifically for each use-case.

For the cracks in concrete, it is too early to discuss structure
similarity. For now, the tessellation based models are visually
convincing and sufficiently versatile for cracks induced by mechanical
stress. A systematic geometric analysis of 3D crack structures,
including  cracks induced by other effects like chemical ones
\cite{ict23} is subject of further research and might lead to more
sophisticated models tailored more closely to the mechanism causing the cracks. 
The concrete example teaches us also that the degree of realism needed heavily 
depends on the used ML model: The RieszNet model is capable of segmenting 
multiscale cracks although having been trained on cracks of one fixed width, only.

The answer for how much of the imaging physics has to be simulated 
correctly seems to depend crucially on both imaging method and 
application: For the SEM, more than the correct electron matter interaction 
seems to be needed. For the CT images of 
concrete samples with cracks on the other hand, no explicit simulation 
of the X-ray imaging is involved at all. Finally, for the optical 
images of structured metal surfaces, surely more parametric models of
surface finishing are needed, while the chosen micro-scale optical properties seem suitable.
However for all other materials beside metal, this is yet to be explored, 
with particular interest on subsurface scattering approximation for transparent 
materials and the use of correct refraction data when simulating glass. 
Finally, to maximize the image appearance similarity, spectral rendering should be 
used instead of classical RGB rendering.
 
\begin{acknowledgement}
This work was supported by the German Federal Ministry of Education and 
Research (BMBF) [grant numbers 05M2020 (DAnoBi), 01IS21058 (SynosIs), 01IS21054 (poST)]. 
We thank Matthias Pahn and Szymon Grezsiak from CE of RPTU for concrete samples and
experimental design, Franz Schreiber from Fraunhofer ITWM and Michael Salamon 
from Fraunhofer EZRT for CT imaging, Markus Kronenberger from
Fraunhofer ITWM for the cubes in Figure 2, Anna Nowacka from
Fraunhofer ITWM and RPTU-CE for the fine-tuned 3d U-net results
featured in Figures~\ref{fig:results-nc} and \ref{fig:results-ppfrc},
and Fraunhofer IOF for machined surface production and topography measurements.
\end{acknowledgement}

\end{document}